\documentclass[sigconf]{acmart}

\usepackage{amsmath,amsfonts}
\usepackage{algorithmic}
\usepackage{graphicx}
\usepackage{textcomp}
\usepackage{xcolor}
\usepackage{booktabs,tabularx}
\usepackage{lipsum}
\usepackage{newtxtext,newtxmath}
\usepackage{subcaption}
\usepackage{caption}
\usepackage{makecell}

\AtBeginDocument{%
  }

\setcopyright{acmlicensed}
\copyrightyear{2025}
\acmYear{2025}
\acmDOI{XXXXXXX.XXXXXXX}

\acmConference[Conference acronym 'XX]{Make sure to enter the correct
  conference title from your rights confirmation emai}{June 03--05,
  2018}{Woodstock, NY}
\acmISBN{978-1-4503-XXXX-X/18/06}

\begin{document}

\title{AS400-DET: Detection using Deep Learning Model \\  for IBM i (AS/400)}

\author{Thanh Tran}
\affiliation{%
 \institution{Amifiable Inc.}
 \city{Meguro City}
 \country{Japan}}
\affiliation{%
  \institution{Japan Advanced Institute of Science and Technology}
  \city{Nomi, Ishikawa}
  \country{Japan}}
\email{thanh.ptit.96@gmail.com}

\author{Son T. Luu}
\orcid{0000-0002-1231-5865}
\affiliation{%
 \institution{Amifiable Inc.}
 \city{Meguro City}
 \country{Japan}}
\affiliation{%
  \institution{Japan Advanced Institute of Science and Technology}
  \city{Nomi, Ishikawa}
  \country{Japan}}
\email{sonlt@jaist.ac.jp}

\author{Quan Bui}
\affiliation{%
 \institution{Amifiable Inc.}
 \city{Meguro City}
 \country{Japan}}
\email{minhquan.bui@amifiable.co.jp}

\author{Shoshin Nomura}
\affiliation{%
  \institution{Amifiable Inc.}
 \city{Meguro City}
 \country{Japan}}
\email{shoshin.nomura@amifiable.co.jp}

\renewcommand{\shortauthors}{Tran et al.}

\begin{abstract}
  This paper proposes a method for automatic GUI component detection for the IBM i system (formerly and still more commonly known as AS/400). We introduce a human-annotated dataset consisting of 1,050 system screen images, in which 381 images are screenshots of IBM i system screens in Japanese. Each image contains multiple components, including \textit{text labels}, \textit{text boxes}, \textit{options}, \textit{tables}, \textit{instructions}, \textit{keyboards}, and \textit{command lines}. We then develop a detection system based on state-of-the-art deep learning models and evaluate different approaches using our dataset. The experimental results demonstrate the effectiveness of our dataset in constructing a system for component detection from GUI screens. By automatically detecting GUI components from the screen, AS400-DET has the potential to perform automated testing on systems that operate via GUI screens.
\end{abstract}




\maketitle

\section{Introduction}
Graphical User Interface (GUI) detection is a fundamental task in software engineering, as it provides an efficient method to automatically identify the components that constitute a system before it encounters errors \cite{kato2024ui}. Detecting components from system screens—such as text boxes, text labels, tables, and keyboards—not only facilitates manipulation but also helps extract relationships among screen components and ensures consistency between the design and implementation phases of system development.

The IBM i system, initially introduced in 1988 as the AS/400, is a robust and versatile operating system developed by IBM. It is designed to support a wide range of business applications and workloads, offering high performance, reliability, and security, which is why this system is still widely used by many companies.
However, since it is written in the legacy programming language COBOL, many organizations are attempting to migrate their systems to Java or other modern languages. During this migration process, given the extensive number of IBM i screen interfaces, manually testing the original specifications of both the legacy system and the migrated system presents significant challenges in terms of program accuracy and consistency. Therefore, we propose a dataset called the IBM i System Object Detection Dataset to facilitate the extraction of GUI components from IBM i system screens, which can aid in testing the screens of the IBM i system. This dataset is a comprehensive and meticulously curated collection of labeled images designed to support the development and evaluation of machine learning models for object detection tasks within the IBM i system interface.

\begin{figure}[H]
    \centering
    \includegraphics[width=\linewidth]{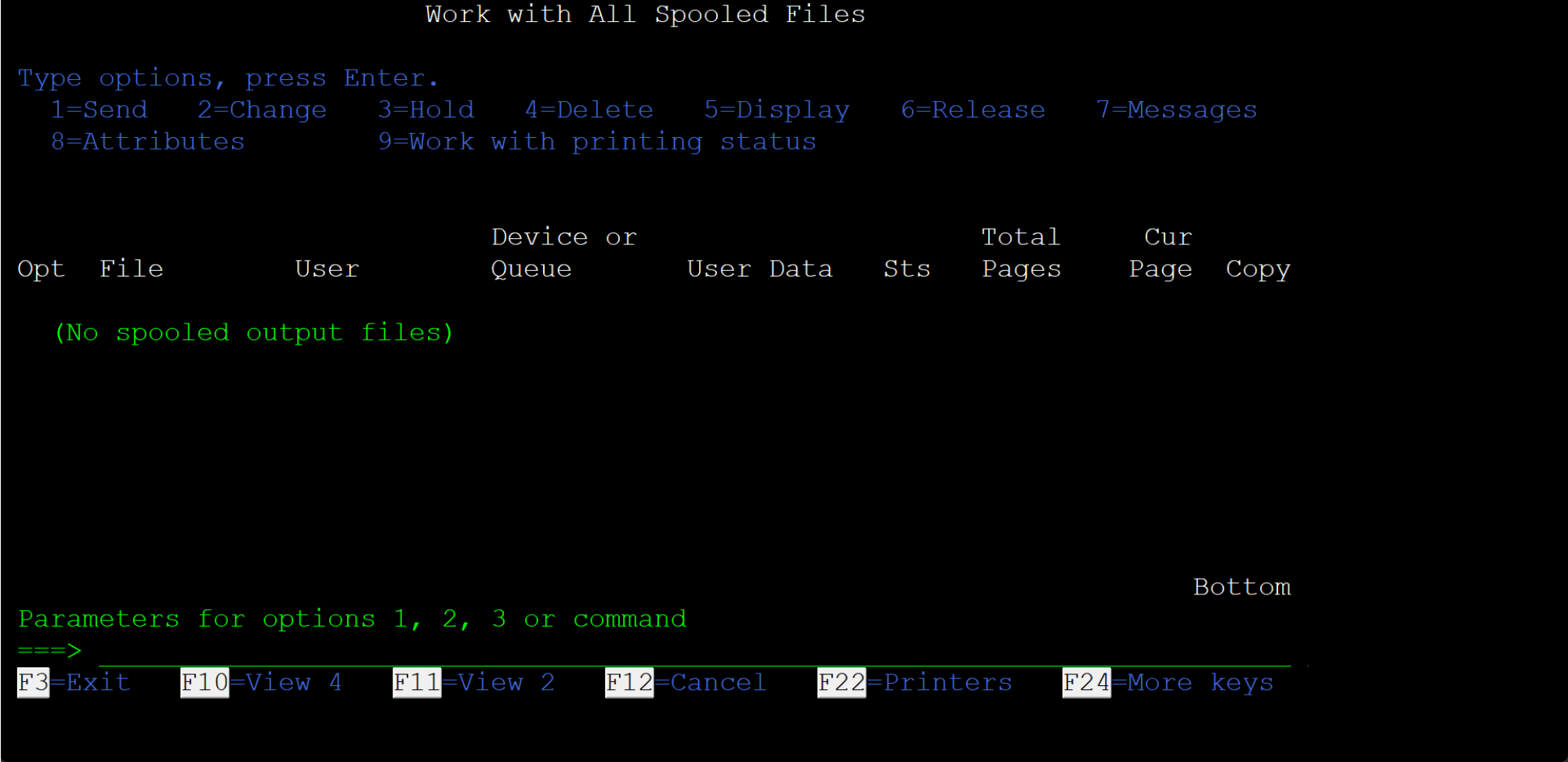}
    \caption{A sample screen from the IBM i system.}
    \label{fig:ibmi_system}
\end{figure}

The dataset consists of a total of 1,050 images of IBM i screens, each containing relevant screen components such as text boxes, text labels, options, table instructions, keyboards, and command lines. Notably, this dataset places a significant emphasis on Japanese-language systems, with 381 of the 1,050 images being screenshots of IBM i system screens in Japanese. Figure \ref{fig:ibmi_system} illustrates a sample image from the IBM i system. The images were collected from a variety of sources, including screenshots of the IBM i system interface under different scenarios and configurations. Each image has been manually annotated using industry-standard tools to ensure accuracy and consistency. The annotations include bounding boxes around objects of interest, along with their respective category labels.

Recent advancements in Optical Character Recognition (OCR) technology and the integration of Large Language Models (LLMs) have significantly improved the accuracy of text recognition \cite{wang2023survey, liu2024ocrbenchhiddenmysteryocr}. However, despite these improvements, accurately obtaining the coordinates of complex GUI components remains a challenge. In such cases, traditional object detection models, such as YOLO and Faster R-CNN, continue to play a crucial role, as they are highly effective in providing precise localization of screen elements, which is essential for tasks like GUI component detection and testing.

The dataset is organized in a structured format to facilitate easy access and utilization. Each image file is accompanied by an annotation file in a widely accepted format (e.g., XML or JSON), which contains the coordinates of the bounding boxes and the labels for each detected object. This structure allows for seamless integration with popular machine learning frameworks and libraries. By providing high-quality annotated images, this dataset aims to advance the field of object detection within the context of the IBM i system, fostering innovation and enhancing the overall user experience. Additionally, we employ an object detection system based on various state-of-the-art deep learning models to evaluate the performance of our proposed dataset.

In summary, this paper has two main contributions. First, we present the IBM i System Object Detection Dataset for extracting components from the GUI screen of the IBM i system. Second, we introduce a deep learning model for automatically extracting components from the GUI screen and evaluate various state-of-the-art object detection models using our dataset. The remainder of the paper is organized as follows. Section \ref{dataset_construction} describes the construction and statistical details of the proposed dataset. Section \ref{experiments} details the object detection experiments conducted on the dataset and the performance of the object detection baselines. Finally, Section \ref{conclusion} concludes the paper.
\section{Related Works}
\label{related_work}
Graphical User Interface (GUI) testing involves verifying the consistency and compliance of GUI components on the screen with specifications and design requirements \cite{yu2023visionbasedmobileappgui}. 
It plays a crucial role in software testing by ensuring that information is correctly presented to simulate actual user actions \cite{8094439}. 
GUI testing encompasses several tasks, as introduced in \cite{yu2023visionbasedmobileappgui}, with GUI Element Detection being a fundamental one \cite{yu2023visionbasedmobileappgui}. 
This task focuses on using vision-based techniques to detect GUI elements on the screen. 
A potential approach to GUI Element Detection is the object detection method. 
Many state-of-the-art deep learning models are available for this task, including convolutional-based models like YOLO \cite{redmon2016lookonceunifiedrealtime}, R-CNN \cite{6909475}, and Faster R-CNN \cite{ren2016faster}, as well as transformer-based models such as Vision Transformer (ViT) \cite{dosovitskiy2021imageworth16x16words} and Detection Transformer (DETR) \cite{detr_resnet}. According to \cite{zaidi2022survey}, the two challenges for object detection in practice include the accuracy of the detection model and the efficiency in resource usage, especially for mobility and lightweight devices. As shown in the works \cite{9532809,wang2023comprehensive}, the YOLO series with higher versions such as YOLOv8 has the potential practical application in real-time detection systems since they can balance between accuracy and execution time. Specifically, the RT-DETR \cite{zhao2024detrsbeatyolosrealtime} with a special hybrid encoder and query selection module performs better than YOLO in both accuracy and execution time for the real-time detection system.

In addition, several previous works in GUI Element Detection have achieved promising results. For example, Kato and Shinozaw \cite{kato2024ui} used DETR to detect UI components from mobile app screens. 
Similarly, the work in \cite{daneshvar2024gui} employed YOLO for detecting UI elements in mobile application screens. 
Specifically, the authors in \cite{yoon2024intent} utilized Large Language Models (LLMs) for GUI testing on mobile application screens. 
Although the method proposed in \cite{yoon2024intent} shows impressive results compared to other baselines, the cost of handling LLMs like GPT-3.5 and GPT-4 is high, posing a challenge for practical deployment. 
While most existing research on GUI screens has primarily focused on mobile application screens, our paper extends this focus to GUI screens of IBM i, which is also widely used in enterprise environments. 
\section{Dataset}
\label{dataset_construction}
\subsection{Dataset Construction}
\begin{figure}[H]
    \centering
    \includegraphics[width=\linewidth]{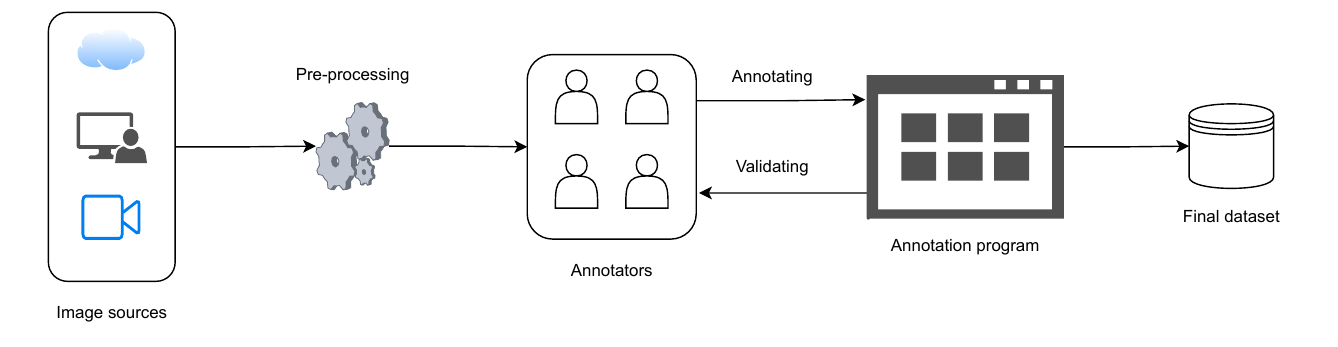}
    \caption{A sample screen from the IBM i system.}
    \label{fig:annotation_process}
\end{figure}
This section describes the process of constructing the AS400-DET dataset.
We began by collecting images of IBM i system screens from various sources, including our internal systems, the internet, and instructional videos about IBM i. We then assembled a group of annotators and used the CVAT software to assist in annotating these images. Figure \ref{fig:annotation_process} provides an overview of our annotation process for building the dataset.

\subsubsection{Dataset Crawling}

We collected screen images of the IBM i system from various sources.
The first source was screenshots from pub400.com\footnote{\url{https://pub400.com/}} system which is a public IBM i server for practice.
To diversify the dataset, we also collected additional images from the web and YouTube videos about the IBM i system, both in Japanese and English. After collecting the images, we manually filtered out those that were irrelevant to the GUI screen.

\subsubsection{Dataset Annotation}
\begin{figure}[H]
    \centering
    \includegraphics[width=\linewidth]{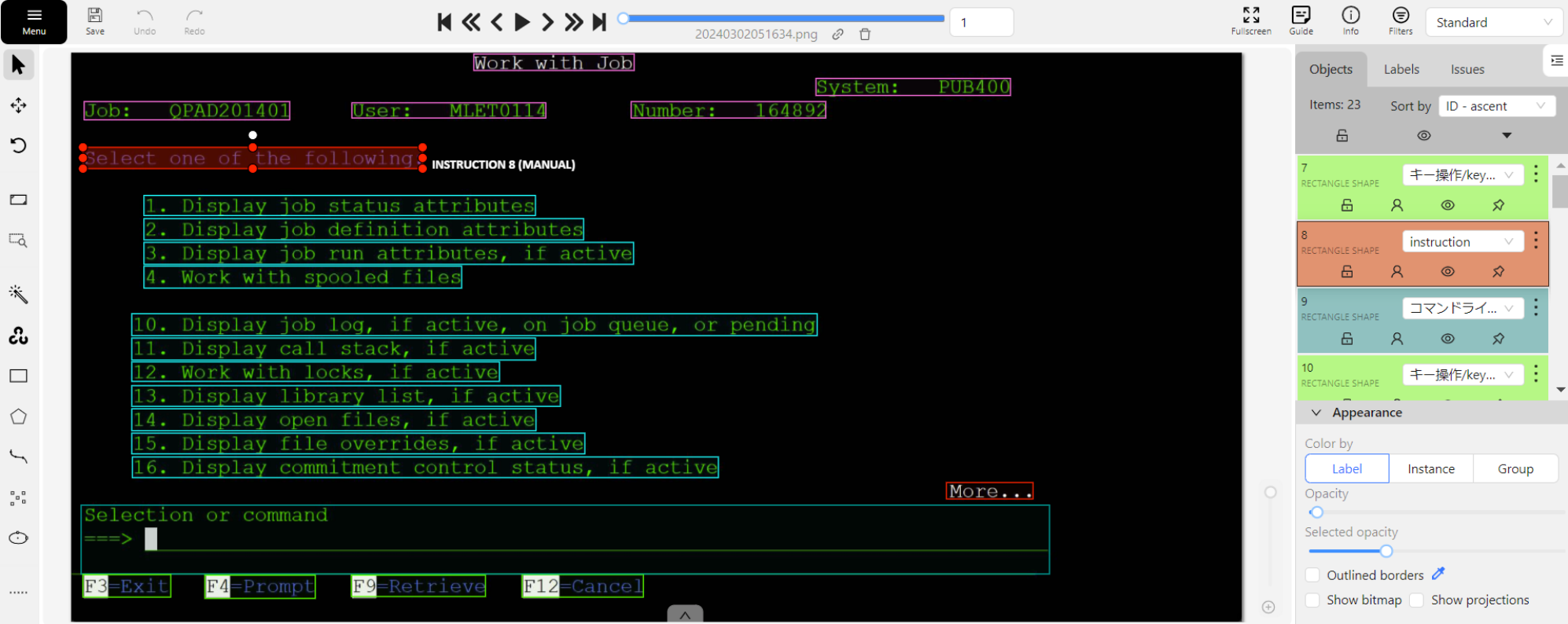}
    \caption{CVAT Annotation Platform.}
    \label{fig:cvat}
\end{figure}

To annotate the collected images, we utilized the CVAT (Computer Vision Annotation Tool)\footnote{\url{https://github.com/cvat-ai/cvat}} - a tool designed for efficient and accurate image and video annotation. The collected images are divided into four parts including Pub400.com, Screenshots, English Website, and Japanese Website. External annotators were hired to use the CVAT interface to manually draw bounding boxes around each object of interest in the images and assign appropriate labels to these boxes. Once the initial annotation is completed, the annotated images will be sent to the quality control process. Senior annotators or supervisors reviewed the annotations to ensure accuracy and consistency. Any identified errors or inconsistencies were corrected before finalizing the annotations. After verification, the annotations were exported in formats compatible with object detection models, such as COCO JSON, Pascal VOC XML, or YOLO text files, ensuring seamless integration with the training pipeline. These annotations play a crucial role in training models to recognize and locate various user-interface elements in different contexts, ultimately leading to more robust and reliable performance in real-world applications. By leveraging CVAT, we ensured the highest quality of annotations, which is essential for the success of both the object detection models and the overall project.
Figure \ref{fig:cvat} illustrates an example annotation based on CVAT software.

\subsection{Dataset Statistics} 
\begin{figure}[H]
    \centering
    \includegraphics[width=\linewidth]{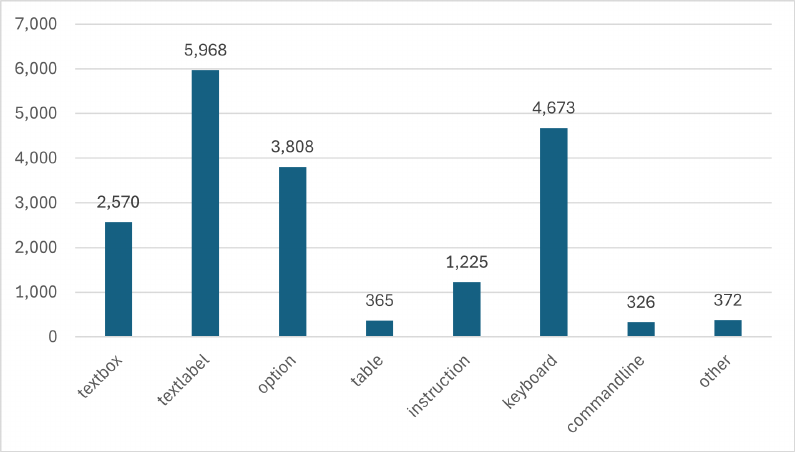}
    \caption{Class distribution in the dataset.}
    \label{fig:class_distribution}
\end{figure}

Table \ref{tab:dataset_stat} provides a summary of the dataset statistics, highlighting the distribution of object classes across its different subsets.
The {IBM i} dataset comprises four distinct subsets: Pub400, Screenshots, English Website, and Japanese Website.
The number of images varies significantly across these subsets, with the English Website containing the most images, followed by the Japanese Website, Pub400, and Screenshots.
As shown in Figure \ref{fig:class_distribution}, the most frequently occurring classes are \textit{textlabel}, \textit{textbox}, and \textit{keyboard}, primarily found in the English Website and Japanese Website subsets. The \textit{option}\footnote{The option includes dot option, equal option, and other options} and \textit{instruction} classes also appear frequently, though to a lesser extent. The least common classes are \textit{commandline} and \textit{table}, indicating fewer instances of these elements in user-interface screenshots. Each subset contributes uniquely to the overall dataset, providing a comprehensive collection of annotated images. The dataset's diverse structure, with varying numbers of images and object class distributions across subsets, ensures broad coverage of user-interface elements.

\begin{table*}[h]
    \centering
    \caption{Dataset statistics}
    \label{tab:dataset_stat}
    \resizebox{.8\textwidth}{!}{
    \begin{tabular}{|l|c|c|c|c|c|c|c|c|c|c|}
     \hline
      & \textbf{\# images} & \textbf{textbox} & \textbf{textlabel} & \textbf{option} & \textbf{table} & \textbf{instruction} & \textbf{keyboard} & \textbf{commandline}  & \textbf{other} \\
     \hline
     Set 1: Pub400 & 229 & 665 & 507 & 621 & 36 & 397 & 1,332 & 68 & 43 \\
     \hline
     Set 2: Screenshots & 57 & 96 & 331 & 152 & 31 & 49 &  267 & 22 & 25 \\
     \hline
     Set 3: English Website & 436 & 1,070 & 3,013 & 1,911 & 190 & 458 & 1,869 & 127 & 188 \\
     \hline
     Set 4: Japanese Website & 328 & 739 & 2,117 & 1124 & 108 & 321 & 1,205 & 109 & 188 \\
     \hline
     Total & 1,050 & 2,570 & 5,968 & 3,808 & 365 & 1,225 & 4,673 & 326 & 372 \\
     \hline
    \end{tabular}
    }
\end{table*}

\begin{figure}[h]
    \centering
    \includegraphics[width=\linewidth]{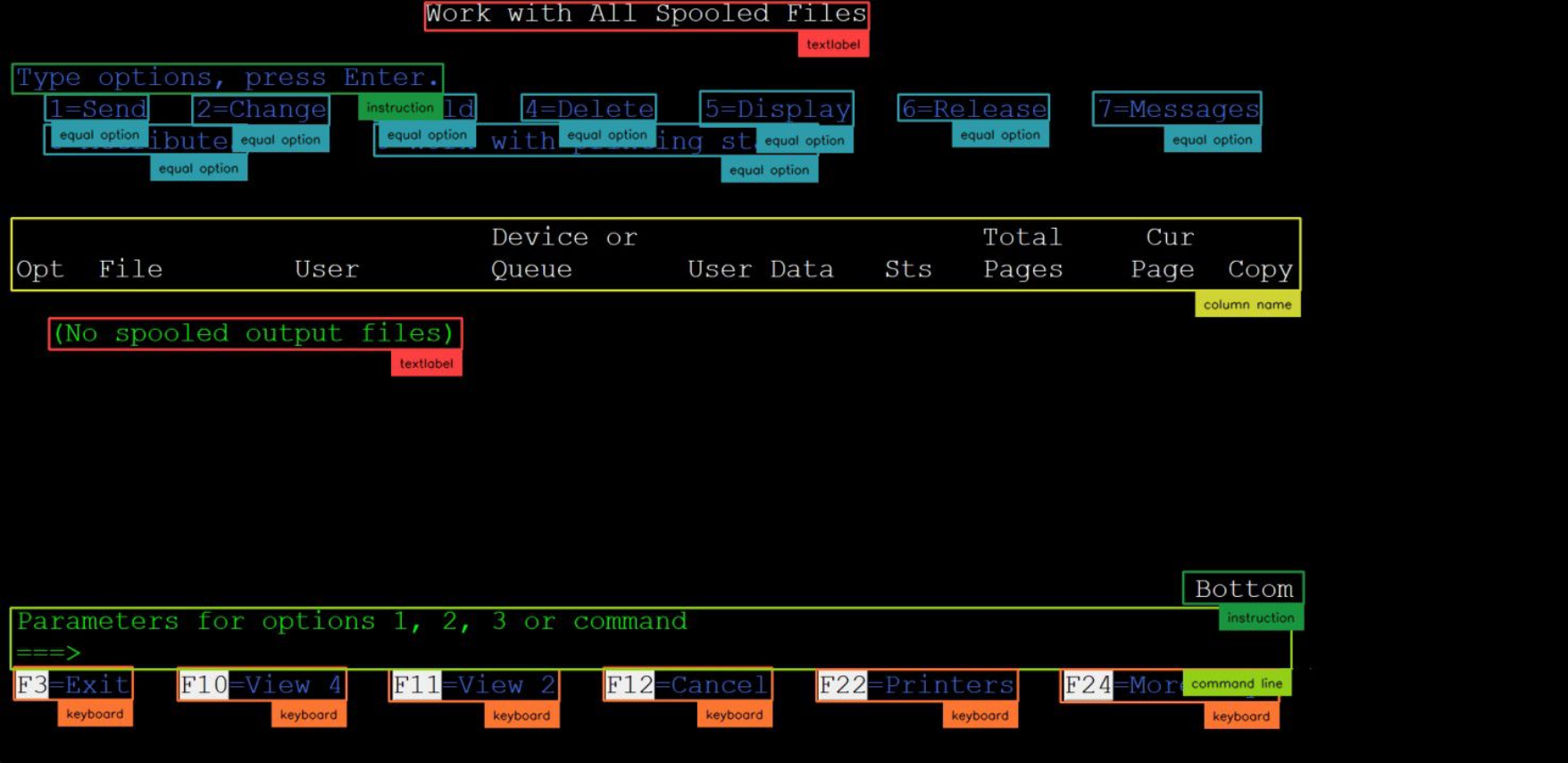}
    \caption{A sample of annotated images.}
    \label{fig:annotated}
\end{figure}

Figure \ref{fig:annotated} illustrates a sample from the dataset with annotations, where each component on the screen is labeled accordingly. These labels enable verifying the adequacy of the displayed information, such as text labels, titles, and instructions, against the design, as well as validating the layout's correctness. Moreover, the extraction of these component labels facilitates automated functional testing by exploiting the position and role of each component on the GUI screen. 

\section{Object Detection Experiments}
\label{experiments}
\subsection{Problem Definition}
The GUI Component detection problem is categorized as the object detection task, which is denoted as:
\begin{itemize}
    \item \textbf{Input}: an image $I$ with the particular size determined by (with, height).
    \item \textbf{Output}: A list of annotation $O = \{o_1, o_2, ... , o_n\}$ determining $n$ components from the image. 
\end{itemize}

For each $o_i \in O$ where $i=1..n$, $o_i = <(x, y, w, h), lb>$, where $(x, y, w, h)$ represents the bounding box of an object in the image, with $x$ and $y$ as the coordinates, and $w$ and $h$ as the width and height of the bounding box, respectively. The label $lb \in \{textlabel, textbox, option, table, instruction, keyboard, commandline\}$ indicates the type of component in the GUI screen.

\subsection{Experiment Preparation} 
The user-interface object detection phase serves as a critical intermediate stage, processing system screenshots to extract relevant elements for use in subsequent applications. To ensure the reliability and consistency of these end-point applications, the object detection stage must deliver high-quality results, accurately identifying and classifying various elements within the input screenshots. Additionally, the object detection method must be resource-efficient, ensuring that the detection process is both fast and cost-effective, which is essential for real-time applications where processing delays can significantly impact usability \cite{9532809,zaidi2022survey}. For the object detection phase, we utilized the following well-established object detection models as baselines:

\begin{itemize} 
    \item DETR-ResNet50 \cite{detr_resnet}: An object detection architecture that integrates a transformer encoder-decoder framework with a ResNet50 backbone, enabling it to model complex relationships between objects in an image. We utilize the \texttt{facebook/detr-resnet-50} checkpoint available on Huggingface. \item YOLOv8X \cite{solawetz2023yolov8}: An advanced object detection model that builds upon the strengths of the YOLO \cite{redmon2016lookonceunifiedrealtime} series. It features an enhanced backbone network that improves feature extraction capabilities compared to previous versions. We employ the checkpoint provided by the Ultralytics framework\footnote{https://github.com/ultralytics/ultralytics}. \item YOLOv9E \cite{wang2024yolov9learningwantlearn}: The latest iteration in the YOLO series, expanding on YOLOv8 with an upgraded backbone, enhanced detection head, and the introduction of dynamic anchor box learning for superior multi-scale detection. We utilize the checkpoint supplied by the author\footnote{https://github.com/WongKinYiu/yolov9}. \item RTDETR-X \cite{zhao2024detrsbeatyolosrealtime}: A state-of-the-art object detection model that merges the advantages of transformer architectures with real-time processing capabilities. It is designed to deliver high detection accuracy while maintaining efficient processing speeds, making it ideal for a variety of real-world applications. We use the checkpoint provided by the author\footnote{https://github.com/lyuwenyu/RT-DETR}. \end{itemize}

\begin{table*}[h]
    \centering
    \caption{Evaluation dataset statistics}
    \label{tab:eval_dataset_stat}
    \resizebox{.8\textwidth}{!}{
    \begin{tabular}{|c|c|c|c|c|c|c|c|c|c|}
     \hline
      & \textbf{\# images} & \textbf{textbox} & \textbf{textlabel} & \textbf{option} & \textbf{table} & \textbf{instruction} & \textbf{keyboard} & \textbf{commandline} & \textbf{other} \\ 
     \hline
     Training set & 838 & 2,011 & 4,614 & 3,029 & 299 & 986 & 3,719 & 261 & 301 \\ 
     \hline
     Validation set & 212 &  559 & 1,160 & 779 & 66 & 239 & 954 & 65 & 71 \\ 
     \hline
    \end{tabular}
    }
\end{table*}


To evaluate the performance of the object detection baselines on the IBM i screenshot dataset, we divided the dataset into training and validation sets. Each subset of the original data was split using an 80/20 ratio, after which the training and validation sets from all subsets were combined. This approach ensures that the dataset is representative and that the models are trained and validated on a diverse range of samples. Table \ref{tab:eval_dataset_stat} presents the statistics of the evaluation dataset. The class distribution in both the training and validation sets is consistent with the distribution observed in each subset. According to Table \ref{tab:eval_dataset_stat},\textit{textlabel}, \textit{option}, and \textit{keyboard} components accounted for a high proportion of the dataset, while the \textit{commandline} and \textit{table} components are few. 


\subsection{Experiment Results}
To evaluate the performance of the object detection models, we use the average precision (mAP) \cite{wang2022parallelimplementationcomputingmean} and average recall (AR) \cite{electronics10030279}. We choose the best checkpoint for each model based on the validation loss and show the empirical results in Table \ref{tab:exp_result}. 
According to Table \ref{tab:exp_result}, the DETR-ResNet50 model shows low performance across all metrics, indicating the struggle to accurately detect and localize objects in the IBM i dataset compared to the other models. In contrast, the YOLOv8X demonstrates robust performance, particularly in terms of mAP@0.5 and mAP@0.75, with a solid AR score, indicating its ability to effectively detect objects with high precision and recall. The improved YOLOv9E model shows slightly better results than YOLOv8X, particularly in mAP@0.5 and mAP@0.75, reflecting a minor enhancement in overall precision and recall. Specifically, the RTDETR-X model outperforms all other models on every metric, boasting notably high mAP scores that indicate excellent precision across various IoU thresholds, along with a high AR score that reflects strong recall. These results suggest that RTDETR-X is particularly effective at both detecting and localizing objects within the evaluation dataset, highlighting its potential for applications in the IBM i system.

\begin{table}[H]
    \centering
    \caption{Experiment Results}
    \label{tab:exp_result}
    \resizebox{.45\textwidth}{!}{
    \begin{tabular}{|c|c|c|c|c|}
     \hline
      & \textbf{mAP@0.5} & \textbf{mAP@0.75} & \textbf{mAP@0.5:0.95} & \textbf{AR(k = 100)} \\ 
     \hline
     DETR-Resnet50 & 27.1 & 8.2 & 11.4 & 24.2 \\ 
     \hline
     YOLOv8X & 75.7 & 69.4 & 57.7 & 63.7 \\
     \hline
     YOLOv9E & 76.8 & 70.5 & 59.0 & 64.8 \\ 
     \hline
     \textbf{RTDETR-X} & \textbf{84.1} & \textbf{80.9} & \textbf{69.3} & \textbf{73.3} \\ 
     \hline
    \end{tabular}
    }
\end{table}

\begin{figure*}[h]
    \centering
    \begin{minipage}{0.32\textwidth}
        \centering
        \includegraphics[scale=0.25]{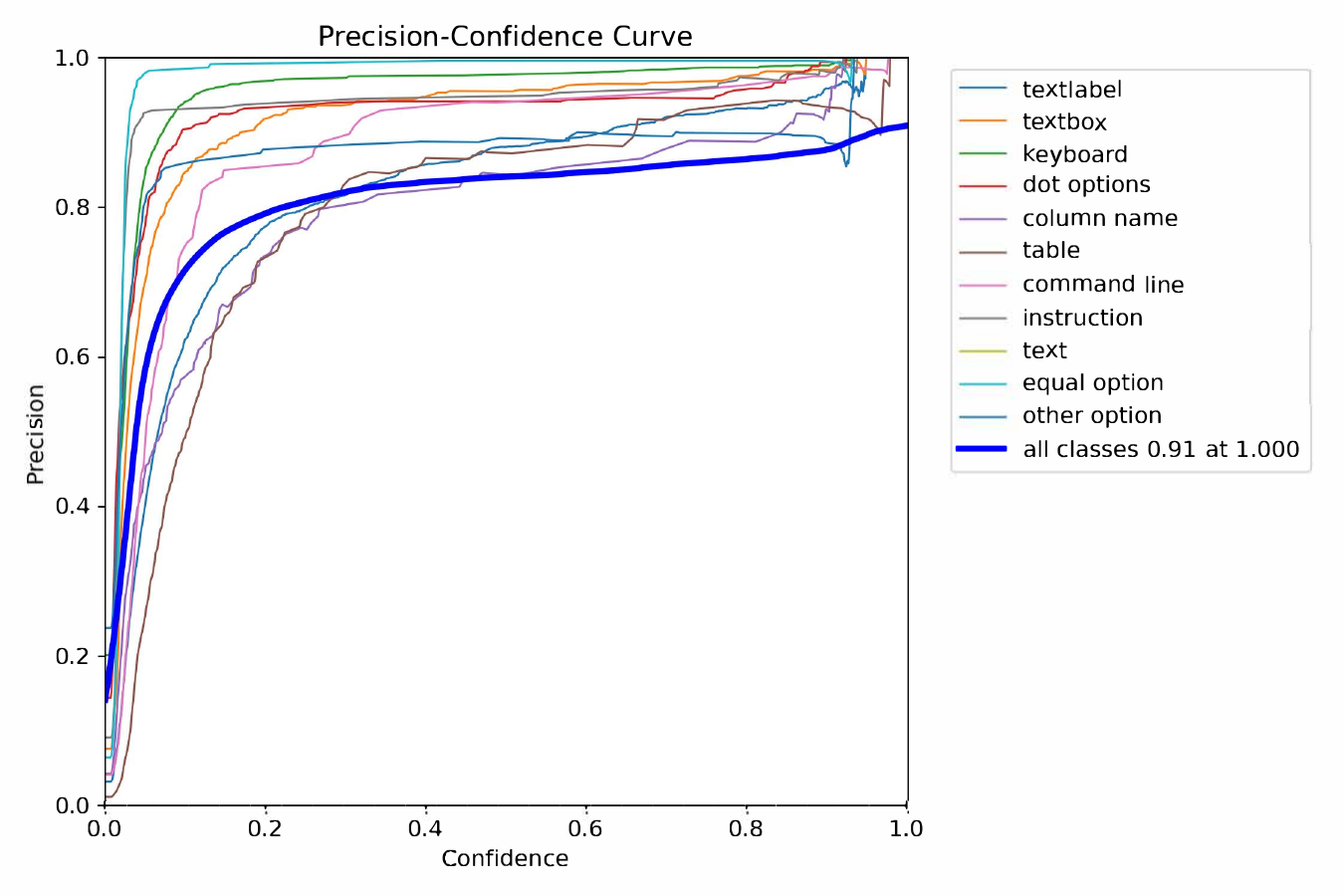}
        \subcaption[first caption.]{}\label{fig:confident_score_precision}
    \end{minipage}
    \begin{minipage}{0.32\textwidth}
        \centering
        \includegraphics[scale=0.25]{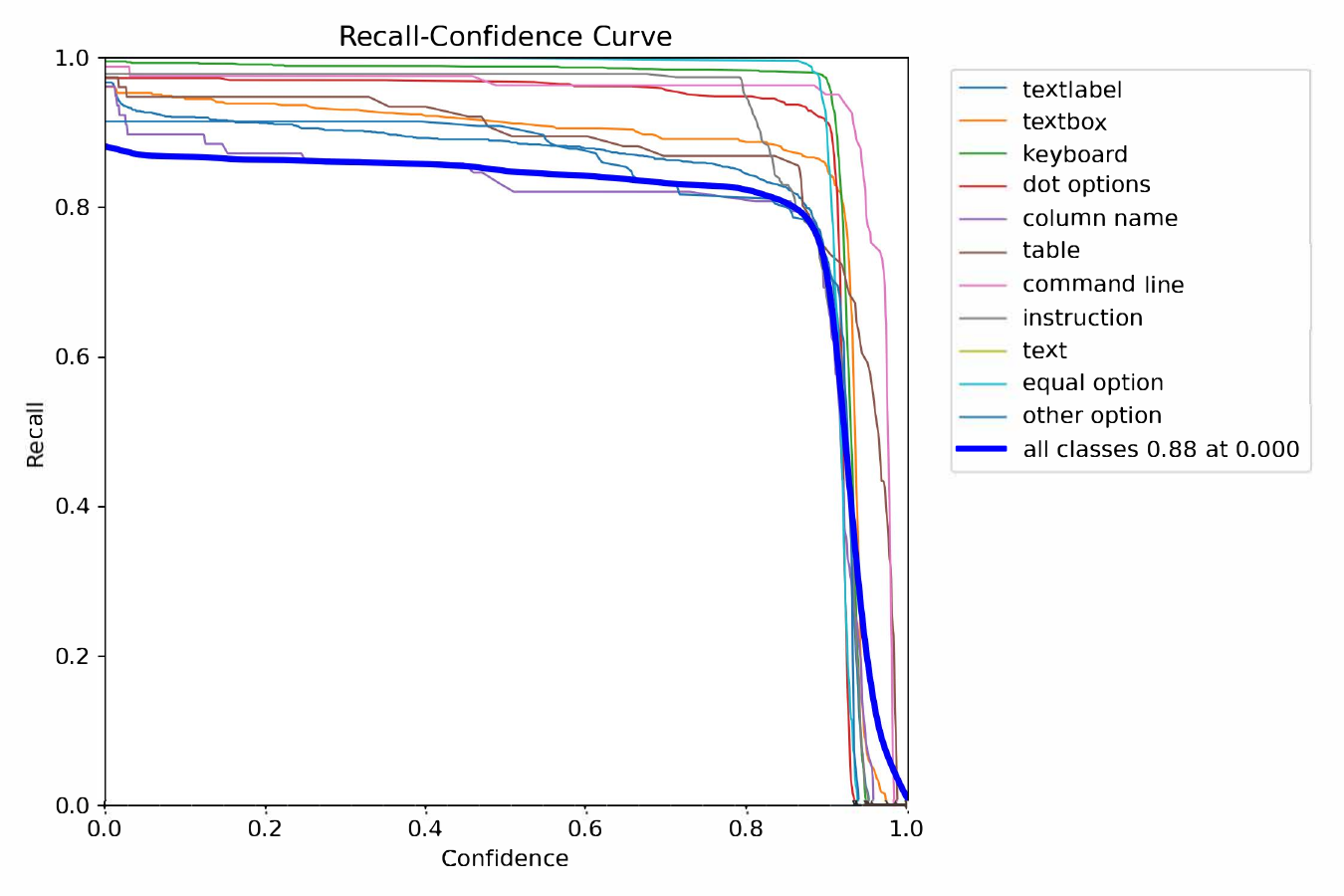}
        \subcaption[second caption.]{}\label{fig:confident_score_recall}
    \end{minipage}
    \begin{minipage}{0.32\textwidth}
        \centering
        \includegraphics[scale=0.25]{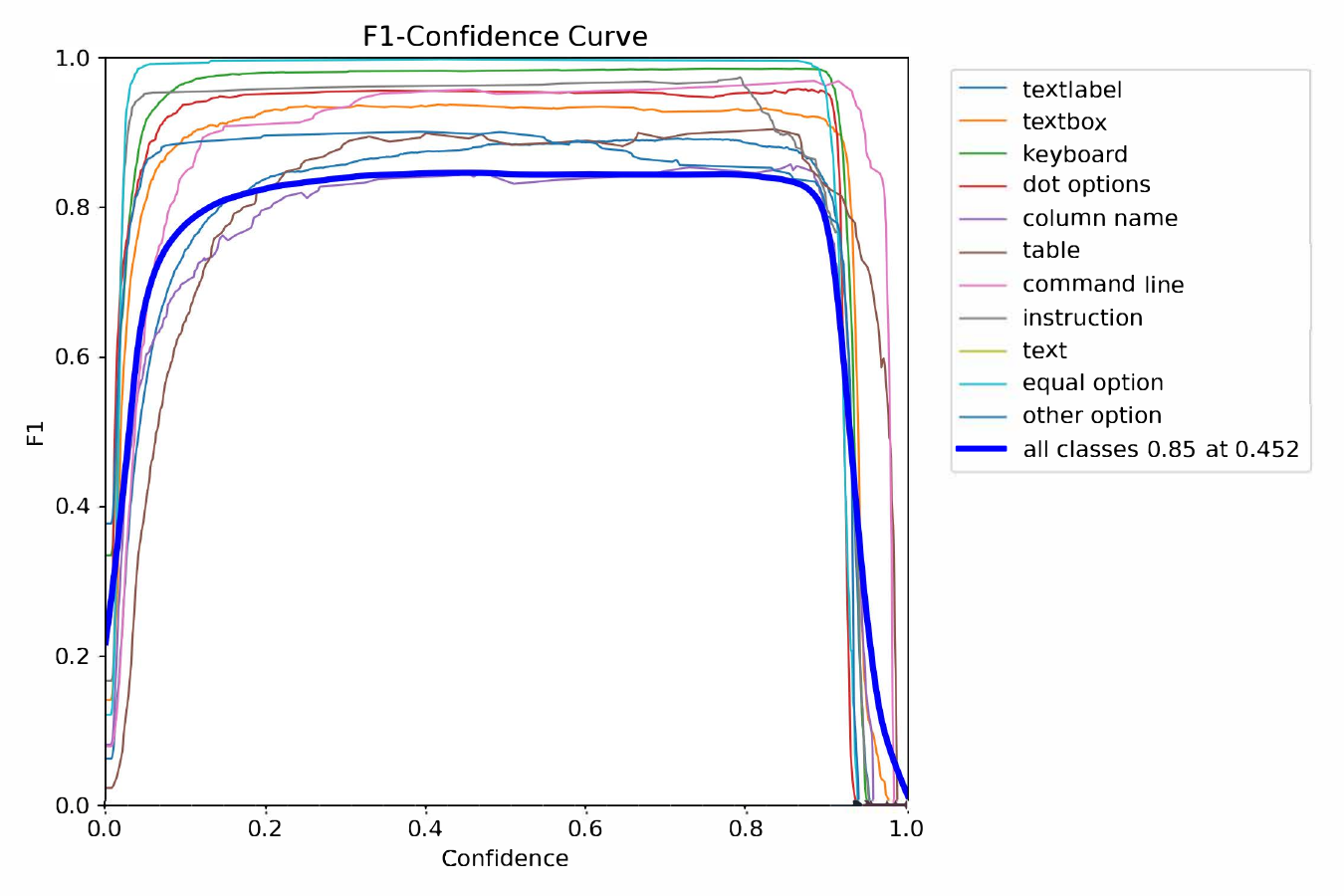}
        \subcaption[third caption.]{}\label{fig:confident_score_f1}
    \end{minipage}
    \caption{The impact of the confidence score to the performance of detection models.}
    \label{fig:confident_score}
\end{figure*}

Additionally, we investigated the impact of the confidence score on the performance of the detection model for each type of component. Figure \ref{fig:confident_score} illustrates how the confidence score affects the accuracy of the RTDETR-X model in detecting component positions within the screen text. The \textit{textlabel} class performs well even at lower confidence levels, while the \textit{table} class requires a higher confidence score to achieve satisfactory results. Notably, when the confidence score exceeds approximately 0.95, the F1 score drops significantly. 
As shown in Figure \ref{fig:confident_score_precision} and \ref{fig:confident_score_recall}, there is a trade-off between precision and recall when adjusting the confidence score. When the confidence score increases, the precision improves proportionally while recall decreases. Overall, we found that a confidence score of 0.452 is optimal for all components. According to Figure \ref{fig:confident_score_f1}, this confidence score allows the detection model to achieve an F1 score of 0.85 across all classes. 

Finally, we conduct the time execution evaluation of the detection model to show the efficiency of the proposed method in real-time screen detection. In our experiment, we measure the time the detecting model needs to perform predicting the screen label for an input image (we ignore the time for loading and reading the images). We randomly choose about 210 samples from the images set (as shown in Table \ref{tab:dataset_stat}) for testing the time of predicting the screen label. Next, we choose the RTDETR-X since it obtains the best performance on the test set of our dataset (as shown in Table \ref{tab:exp_result}) to measure the prediction time. We run the evaluation on a computer with 16Gb RAM, CPU Intel Core i7-1360P (16 GPUs, 2.2GHz), and 500GB SSD hard disk drive. 

\begin{table}[h]
    \centering
    \caption{Statistical information about the prediction time of the detection model}
    \label{tab_time_exec_stat}
    \begin{tabular}{lr}
    \hline  
        & \textbf{Prediction time (second)} \\
        \hline
        \textbf{Mean}    & 0.6304             \\
        \textbf{Median} & 0.6314             \\
        \textbf{Max}     & 0.8606             \\
        \textbf{Min}     & 0.4149             \\
        \textbf{Std.}    & 0.0538             \\
        \hline
    \end{tabular}
\end{table}

\begin{figure}[H]
    \centering
    \includegraphics[width=\linewidth]{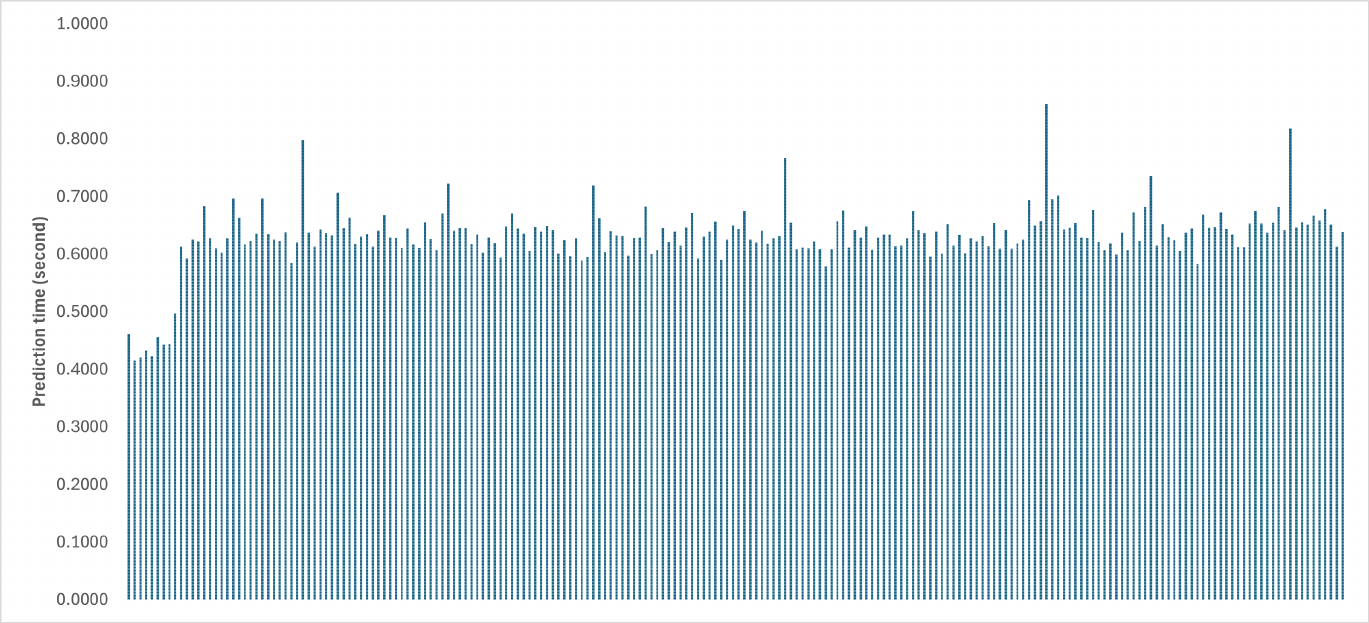}
    \caption{Time execution (second) of detection model on 210 samples.}
    \label{fig:time_exec_210_samples}
\end{figure}

As shown in Table \ref{tab_time_exec_stat}, a sample takes approximately 0.63 seconds to predict the screen label. In some cases, the detection model may take up to 0.8 seconds to perform prediction.  Figure \ref{fig:time_exec_210_samples} illustrates the prediction time for 210 test samples by our detection model. The difference between the prediction time for each image is about 0.0538 second. In general, the detection model requires less than 1 second for prediction, indicating that it is efficient enough to integrate into real-time systems.

\subsection{Results Analysis} 
\begin{figure}[ht]
    \centering
    \includegraphics[width=1.1\linewidth]{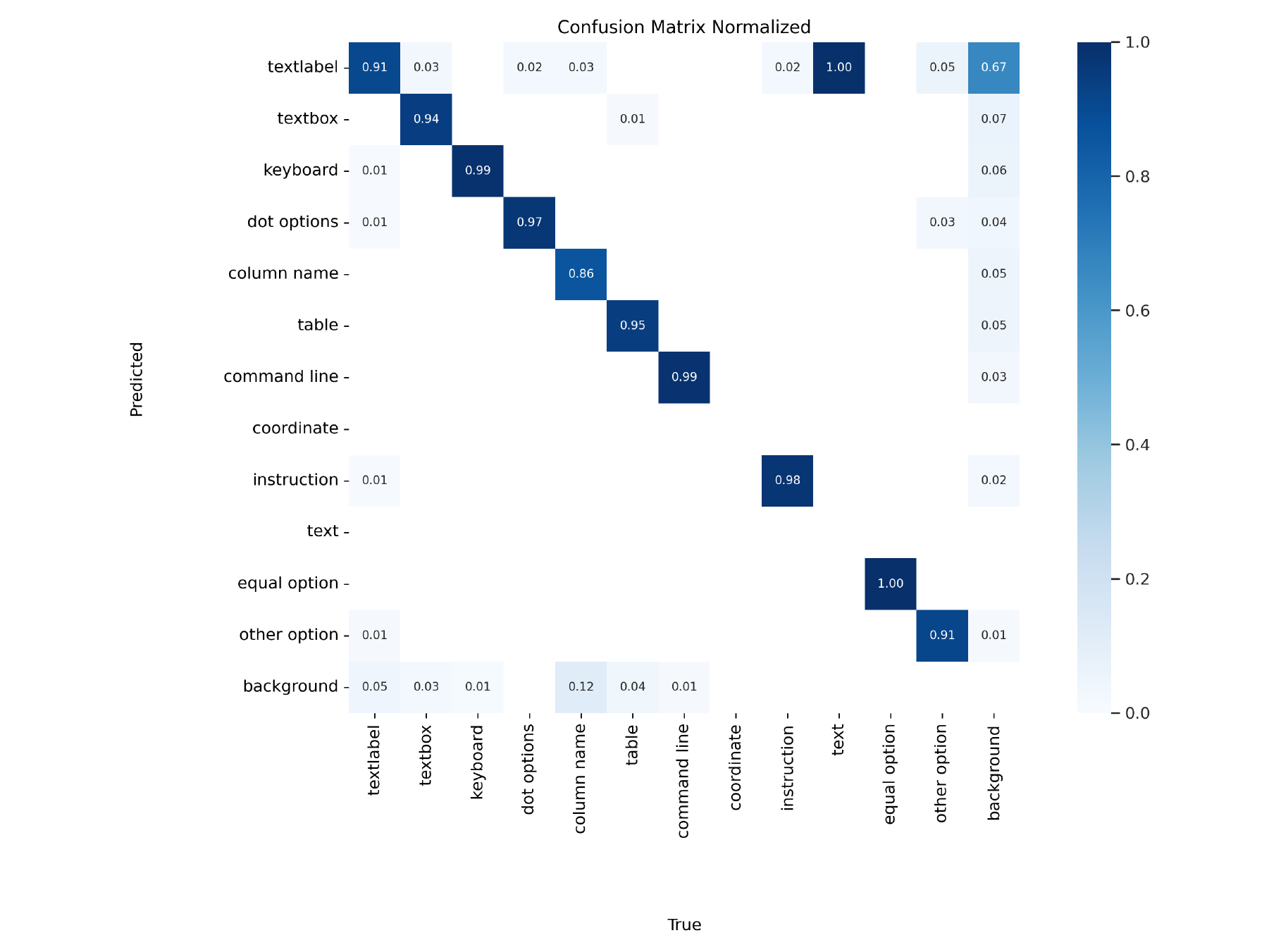}
    \caption{Confusion matrix of detection model for each screen component.}
    \label{fig:confusion}
\end{figure}

\begin{figure*}[ht]
    \centering
    \resizebox{\textwidth}{!}{
    \begin{minipage}{.48\textwidth}
        \centering
        \includegraphics[width=\textwidth]{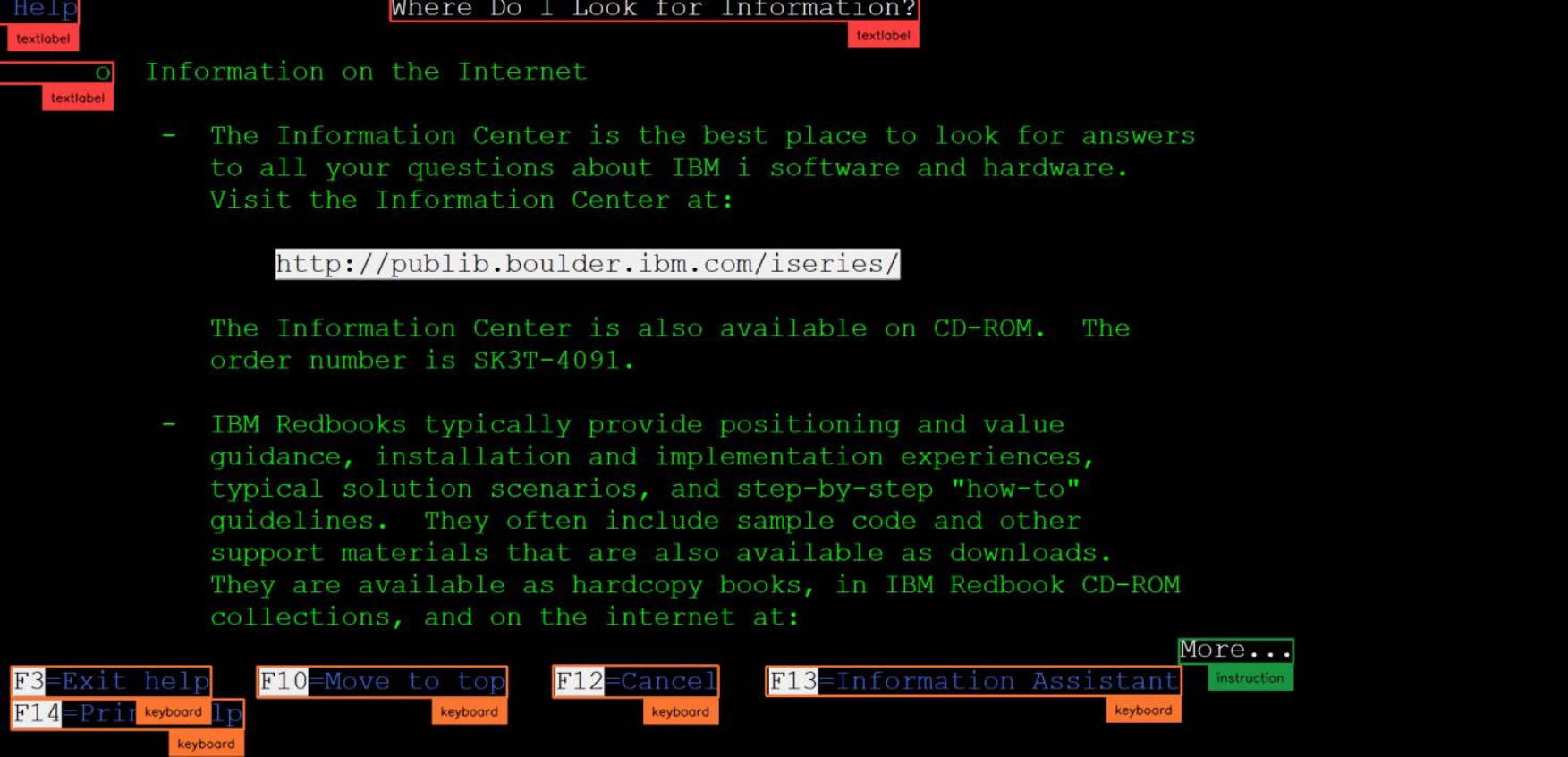}
        \subcaption[first caption.]{Predicted}
        \label{fig:error1_a}
    \end{minipage}
    \begin{minipage}{.48\textwidth}
        \centering
        \includegraphics[width=\textwidth]{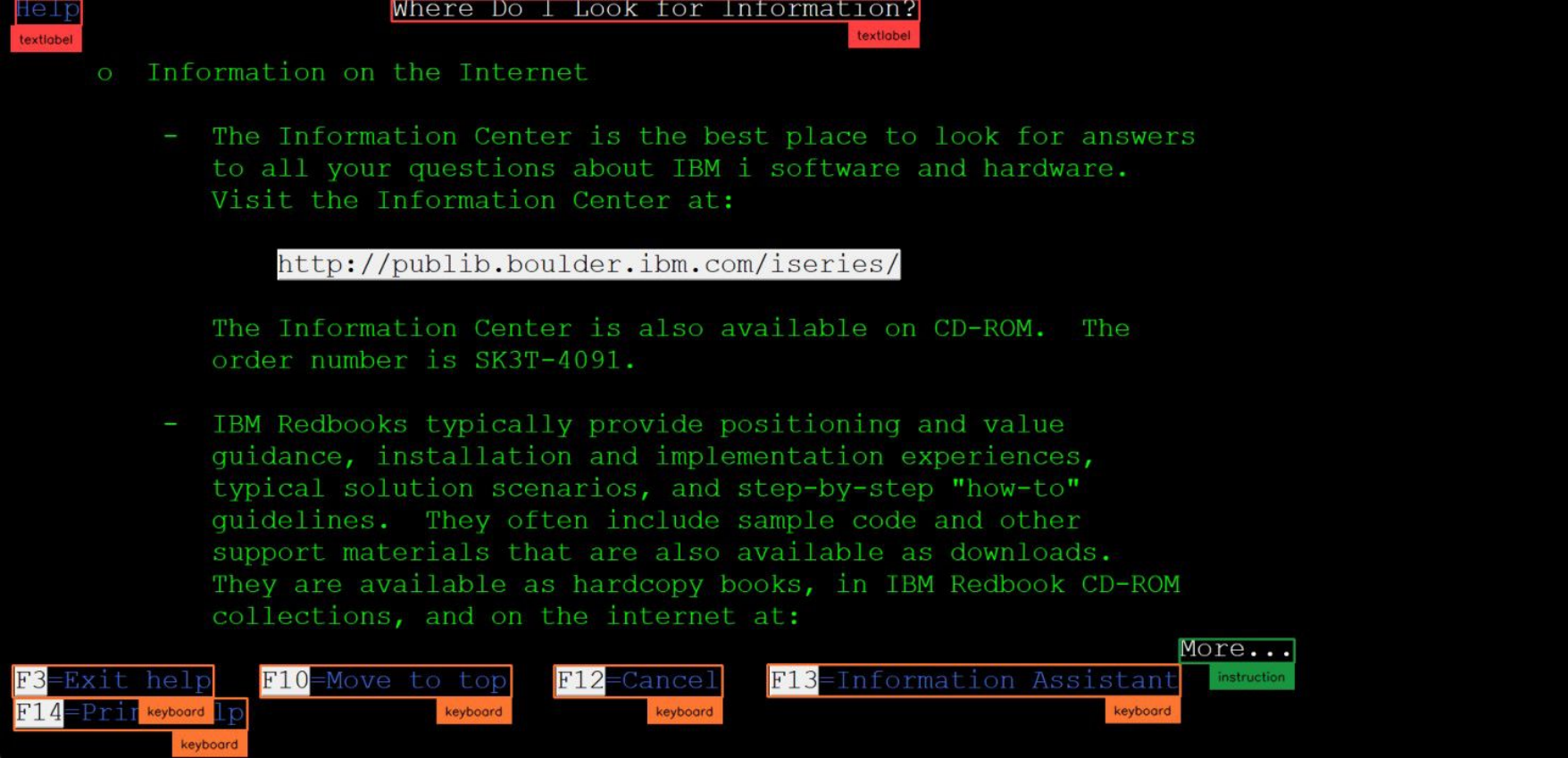}
        \subcaption[second caption.]{Ground truth}
        \label{fig:error1_b}
    \end{minipage}
    }
    \caption{Prediction sample 1.}
    \label{fig:error1}
\end{figure*}

\begin{figure*}[ht]
    \centering
    \resizebox{\textwidth}{!}{
    \begin{minipage}{.48\textwidth}
        \centering
        \includegraphics[width=\textwidth]{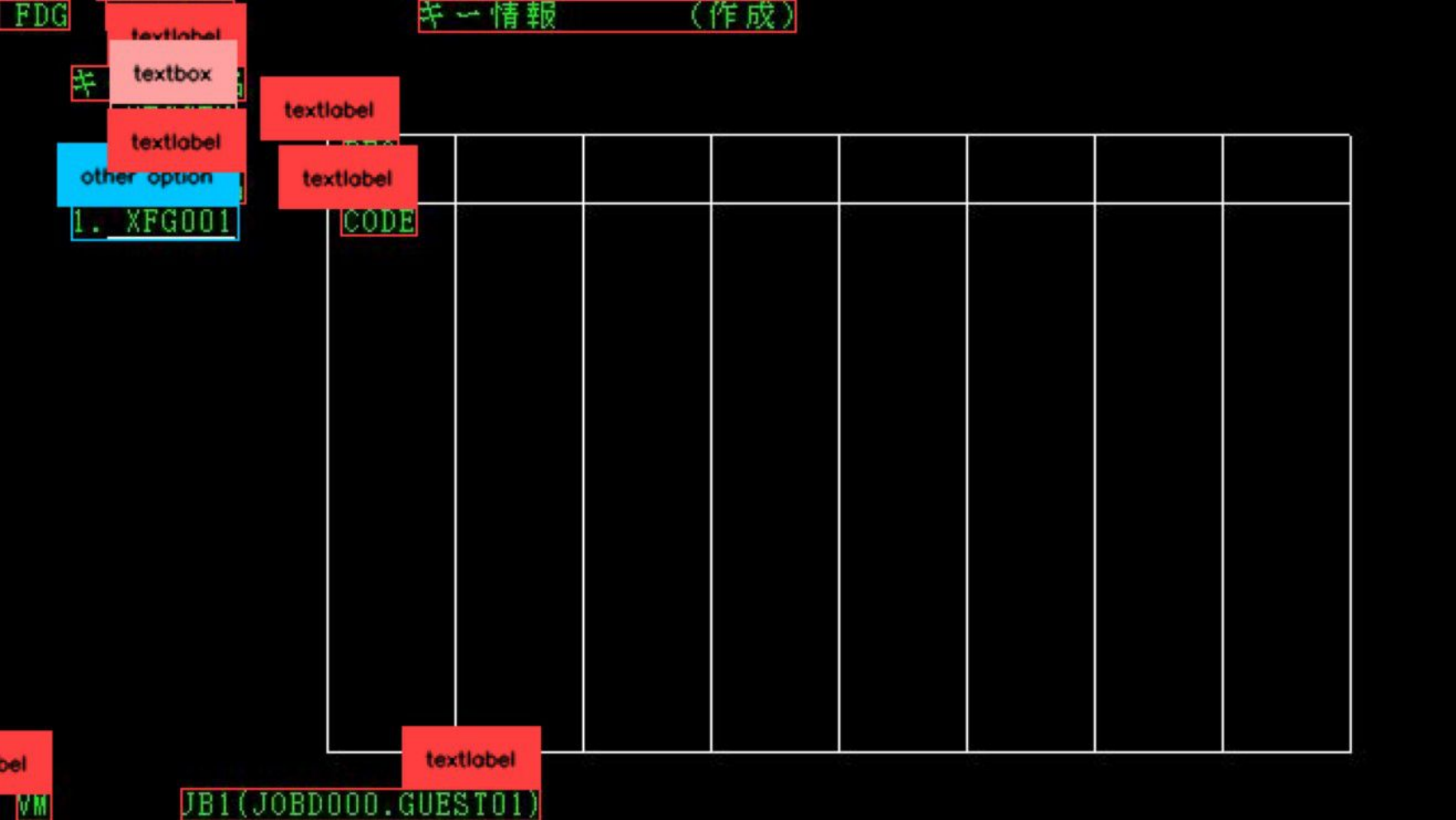}
        \subcaption[first caption.]{Predicted}
        \label{fig:error2_a}
    \end{minipage}
    \begin{minipage}{.48\textwidth}
        \centering
        \includegraphics[width=\textwidth]{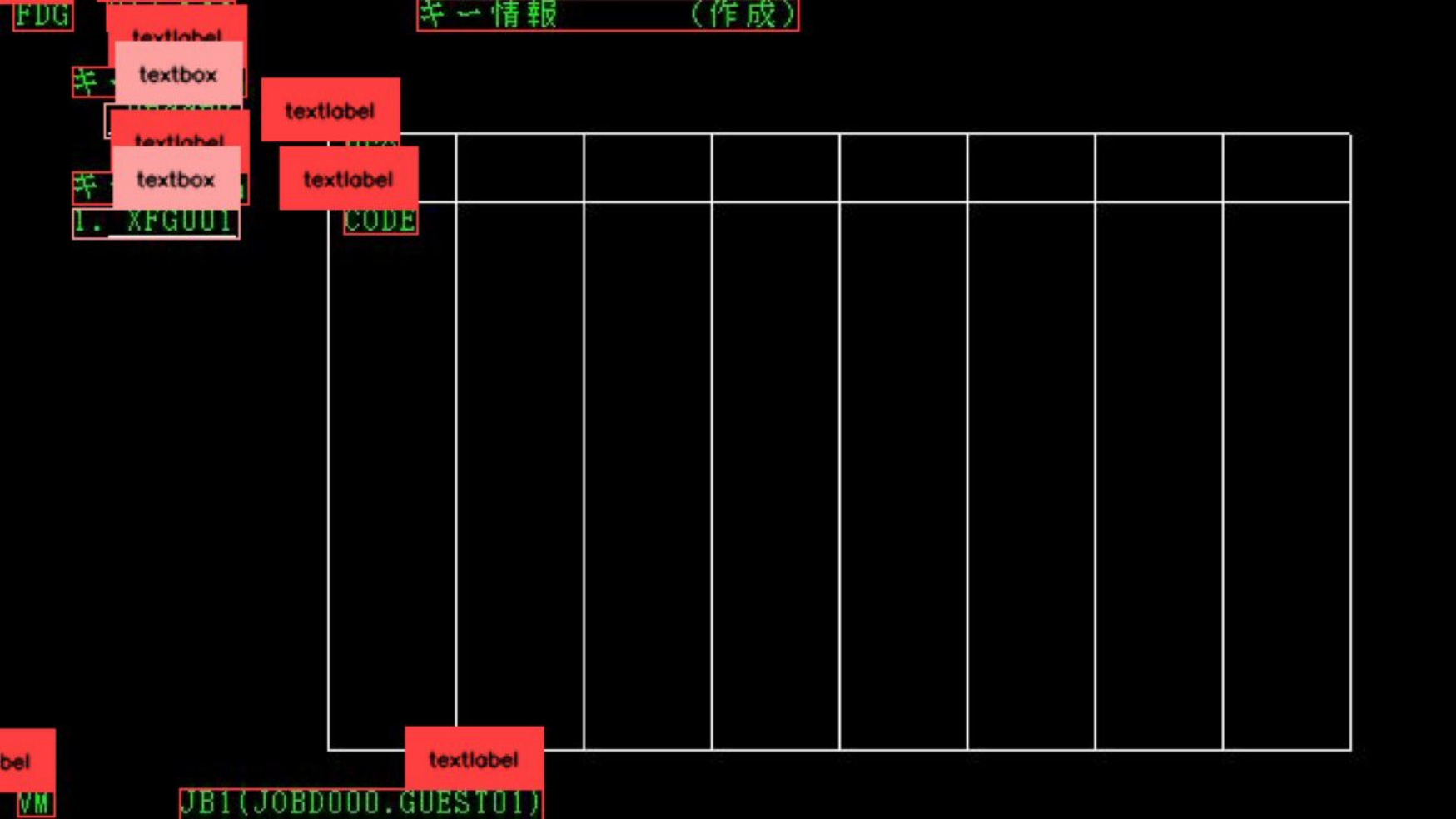}
        \subcaption[second caption.]{Ground truth}
        \label{fig:error2_b}
    \end{minipage}
    }
    \caption{Prediction sample 2.}
    \label{fig:error2}
\end{figure*}

\begin{figure*}[ht]
    \centering
    \resizebox{\textwidth}{!}{
    \begin{minipage}{.48\textwidth}
        \centering
        \includegraphics[width=\textwidth]{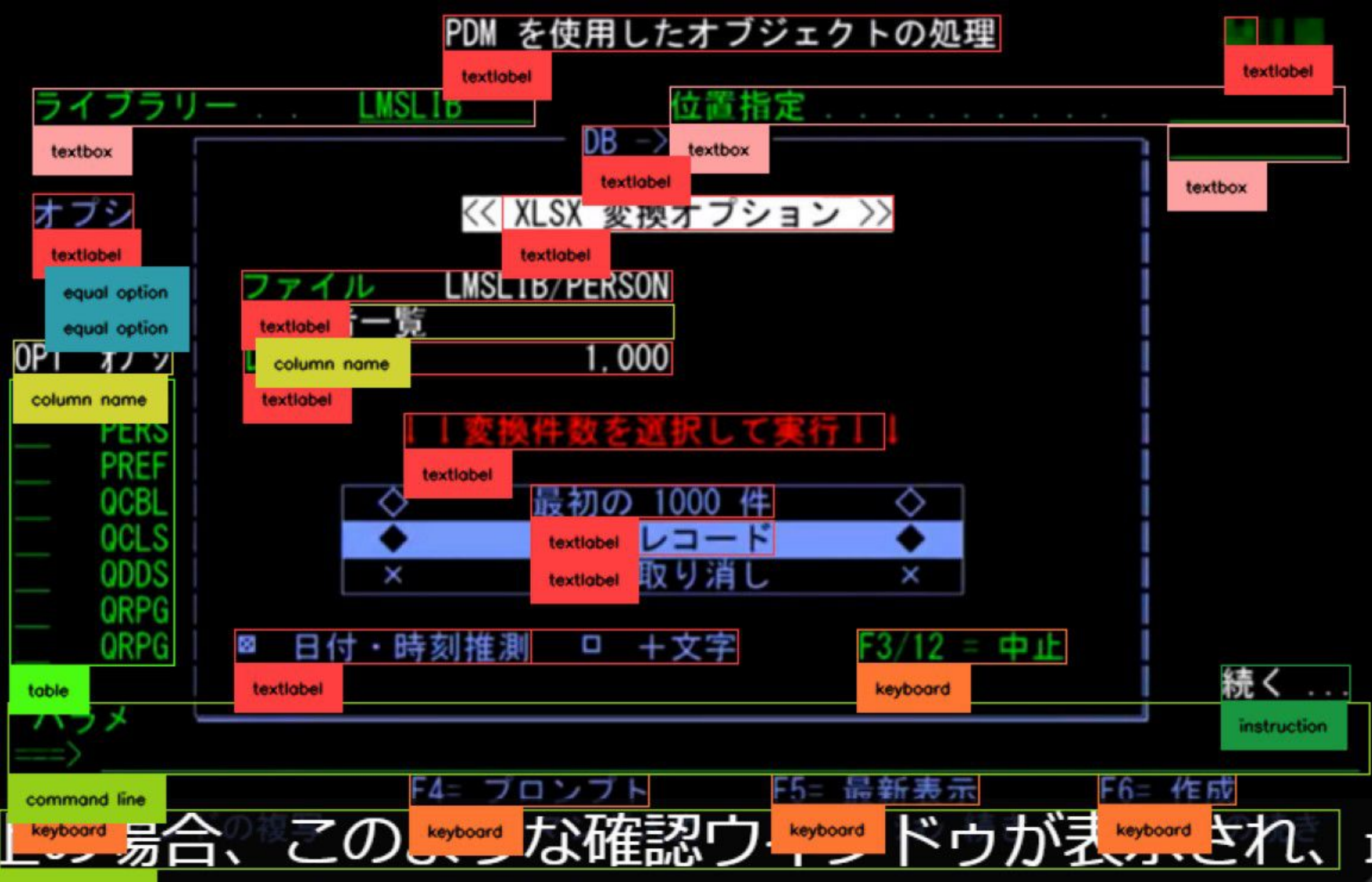}
        \subcaption[first caption.]{Predicted}
        \label{fig:error3_a}
    \end{minipage}
    \begin{minipage}{.48\textwidth}
        \centering
        \includegraphics[width=\textwidth]{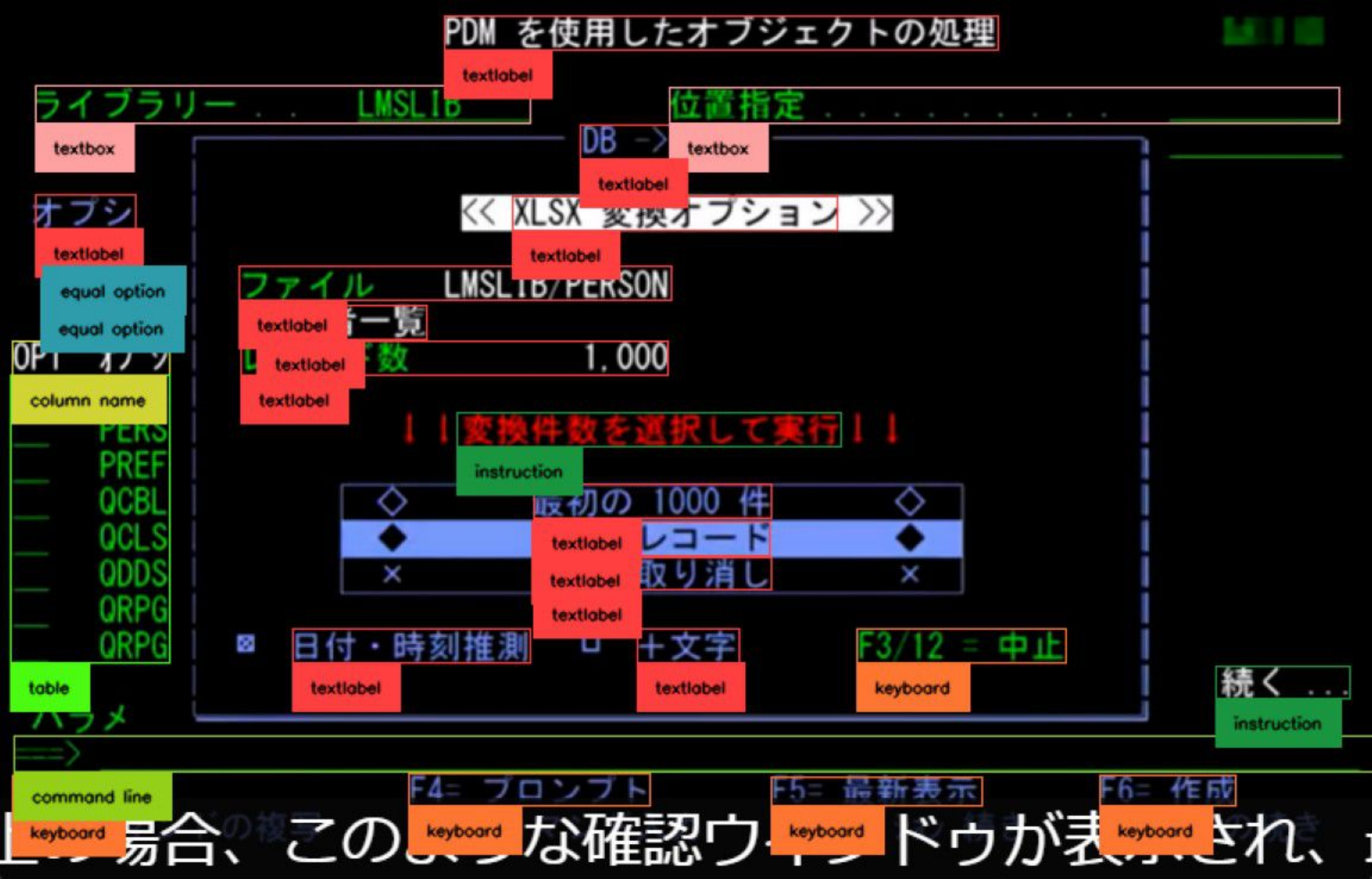}
        \subcaption[second caption.]{Ground truth}
        \label{fig:error3_b}
    \end{minipage}
    }
    \caption{Prediction sample 3.}
    \label{fig:error3}
\end{figure*}

\begin{figure*}[ht]
    \centering
    \resizebox{\textwidth}{!}{
    \begin{minipage}{.48\textwidth}
        \centering
        \includegraphics[width=\textwidth]{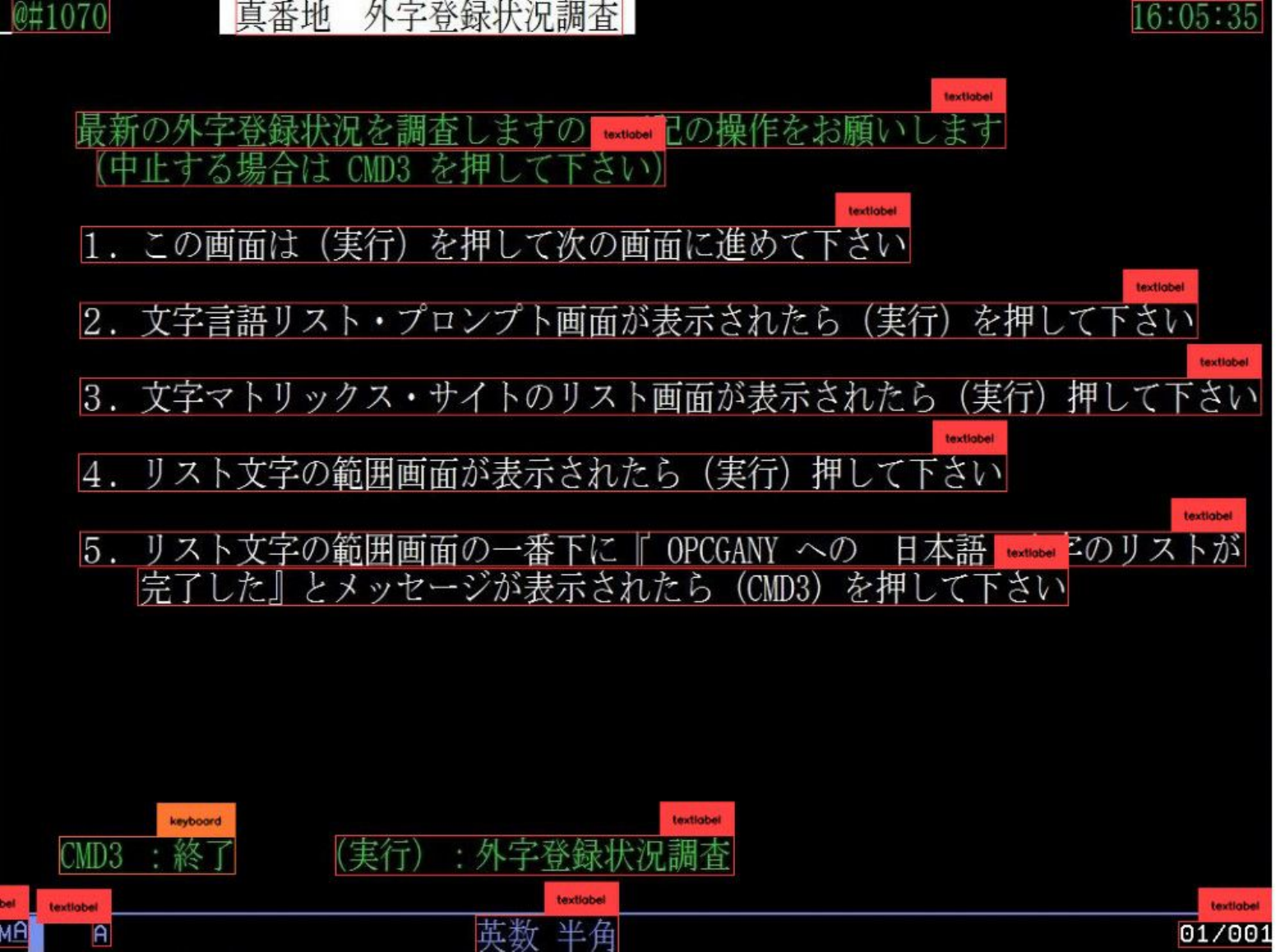}
        \subcaption[first caption.]{Predicted}
        \label{fig:error4_a}
    \end{minipage}
    \begin{minipage}{.48\textwidth}
        \centering
        \includegraphics[width=\textwidth]{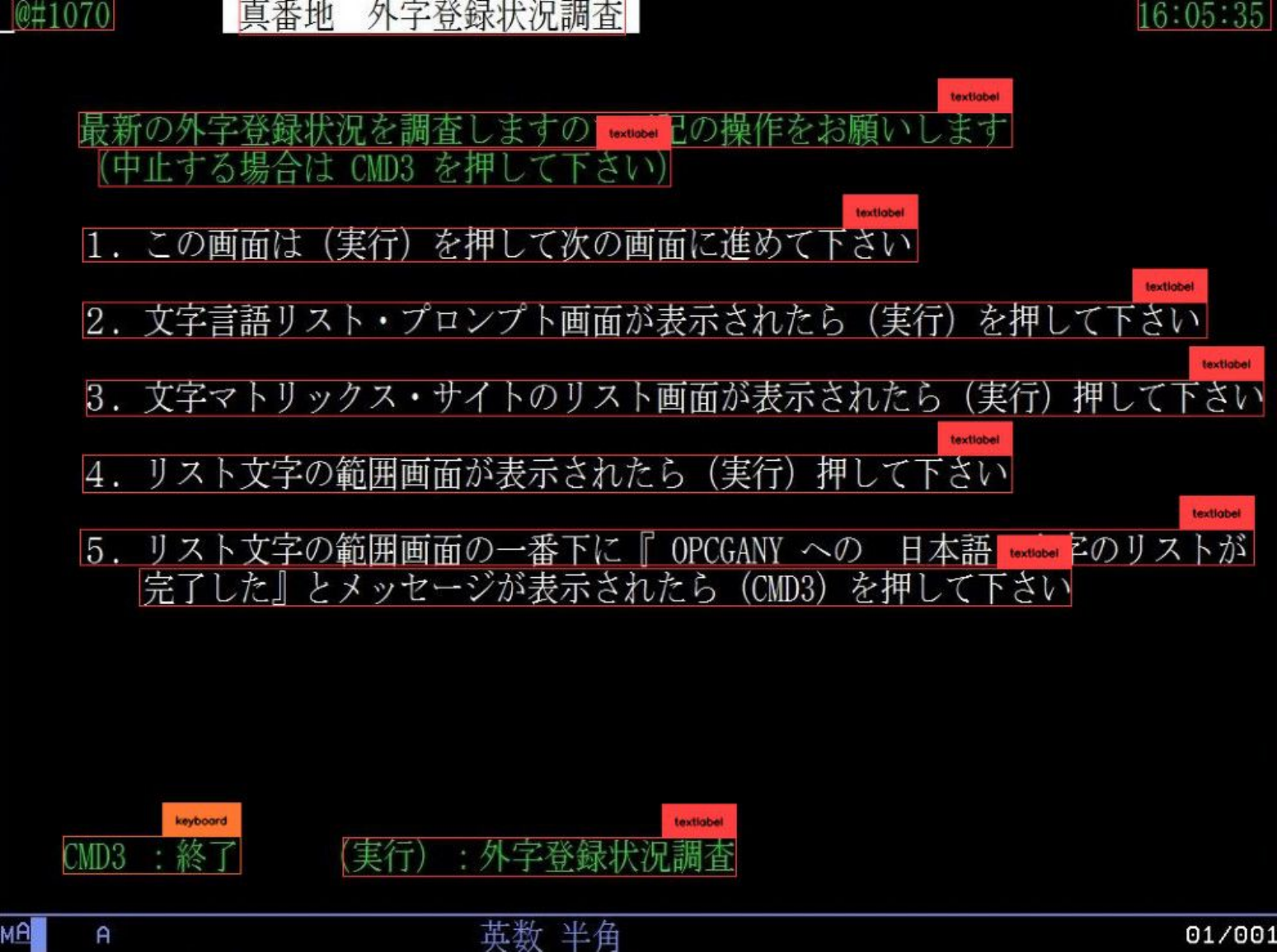}
        \subcaption[second caption.]{Ground truth}
        \label{fig:error4_b}
    \end{minipage}
    }
    \caption{Prediction sample 4.}
    \label{fig:error4}
\end{figure*}

In this section, we analyze the errors based on the predictions made by the RTDETR-X model. According to Figure \ref{fig:confusion}, the model accurately identifies most screen components. However, the \textit{textlabels} class is often misclassified, frequently being mistaken for \textit{text} or \textit{background}. Additionally, the \textit{background} is sometimes incorrectly identified as the \textit{column name} label on the screen.

We also reviewed several prediction examples, as shown in Figures  \ref{fig:error1}, \ref{fig:error2}, \ref{fig:error3}, and \ref{fig:error4}.
In Figure \ref{fig:error1}, the model incorrectly identified a "dot" symbol on the GUI screen as a \textit{textlabel} (see the top left of Figure \ref{fig:error1_a}) due to its visual similarity to the letter "O." In Figure \ref{fig:error2}, the option "1. XFG001" on the middle left of image  \ref{fig:error2_a} was predicted as "other option" rather than "textbox" (see image  \ref{fig:error2_b}), likely because it resembles a GUI screen with numeric choices and descriptive text.

In Figure  \ref{fig:error3}, two misclassifications were observed. The first occurred in the middle of the screen, where the text below "LMSLIB/PERSON" was mistakenly detected as a column name, possibly due to its arrangement in a way that resembles a table format (as shown in image \ref{fig:error3_a}). The second error involved a line at the top right of the screen (see image  \ref{fig:error3_a}) that was classified as a \textit{textbox}, likely because it resembled an input placeholder. In screens, underscore lines are commonly used as input placeholders, but they are usually centered. In this instance, the line was merely a background detail, leading to the model's confusion.

Similarly, in Figure  \ref{fig:error4}, the detection model incorrectly labeled the footer of the screen (see image \ref{fig:error4_a}) as a \textit{textlabel}, when in reality, the footer was simply a background element (see image  \ref{fig:error4_b}). Although the \textit{textlabel} takes a large proportion in the dataset (as shown in Figure \ref{fig:class_distribution}), the model is still misclassified in this label, indicating the need for improvement in the ability of the detection model. From these findings, we identified two key challenges contributing to misdetections: the unusual positioning of \textit{textlabels} on the screen and the model's limited contextual understanding of the \textit{textlabel}. 

Overall, our detection system is shown to be efficient in detecting screen components, particularly common components such as \textit{text labels, text boxes, options, and keyboards}. To further improve performance, we need to integrate contextual and positional information into the detection model to enhance its ability to detect components in more challenging or atypical cases.

\section{Conclusion}
\label{conclusion}
In this paper, we introduced AS400-DET, a method for automatically detecting GUI components from IBM i screens. We began by presenting a human-annotated dataset specifically designed for GUI component detection, consisting of 1,050 images, including 381 images from Japanese systems. Each image contains multiple components, such as \textit{textbox}, \textit{textlabel}, \textit{option}, \textit{instruction}, \textit{keyboard}, \textit{table}, and \textit{command line}.

Besides, we developed a detection system based on deep learning models to identify these components on the screen accurately, and we evaluated several robust image segmentation models using our dataset. Among the models tested, RTDETR-X achieved the highest performance, with a mAP@0.5 score of 84.1, demonstrating the effectiveness of our detection system for practical applications. Our error analysis revealed that the current detection model lacks a comprehensive understanding of positional and contextual information regarding the text on the screen. Besides, our detection system takes less than 1 second to detect the screen labels, which is efficient enough to deploy in practice.

In future work, we plan to deploy our proposed method to end users and enhance its capability to interpret position-aware and contextualized text on the screen. We aim to achieve this by leveraging advanced transformer models like vision-language models such as CLIP \cite{radford2021learning} and ViLT \cite{kim2021vilt} since these models can capture both visual and textual information from the screen, which will improve the accuracy of the detection system.

\begin{acks}
This work is supported by Amifiable Inc. We would like to express our thanks to the annotators for their valuable contribution in constructing the dataset.
\end{acks}
\bibliographystyle{ACM-Reference-Format}
\bibliography{references}

\end{document}